\documentclass[]{fairmeta}
\AtBeginDocument{%
  }

\usepackage{paralist}

\title{Evaluating the Quality of Randomness and Entropy in Tasks Supported by Large Language Models}

\author[1\dagger]{Rabimba Karanjai}
\author[1]{Yang Lu}
\author[2]{Ranjith Chodavarapu}
\author[2]{Lei Xu}
\author[1]{Weidong Shi}

\affiliation[1]{University Of Houston}
\affiliation[2]{Kent State University}
\contribution[\dagger]{rkaranjai@uh.edu}

\abstract{The rapid advancement of large language model (LLM) technology has led to diverse applications, many of which inherently require randomness, such as stochastic decision-making, gaming, scheduling,  AI agents,  and cryptography-related tasks. However, the capabilities of LLMs in handling randomness, particularly in generating and utilizing random numbers effectively, remain unclear. This paper investigates the capacity of LLMs for handling tasks that involve randomness through a series of experiments. We designed a set of experiments that consider various factors that can influence an LLM's performance in tasks involving randomness, such as accessibility to external tools, types of tasks, model states (fresh vs. non-fresh), and prompting strategies. The experiments cover a range of tasks, including generating random numbers, generating random strings such as passwords, shuffling items, and evaluating the quality of randomness using entropy and the NIST randomness test-suite. Our findings reveal that while LLMs can generate outputs that exhibit some degree of randomness, their performance is inconsistent and often deviates significantly from the expected behavior. The analysis of the experimental results highlights key limitations and areas where improvement is needed for the LLMs to effectively handle tasks involving randomness. }

\date{04 August 2025} 

\begin{document}
\maketitle

\section{Introduction}


Large language models (LLMs~\cite{touvron2023llama}) have found numerous applications, such as
natural language processing (NLP), machine translation, source code generation and translation,  question-answering, chatbots~\cite{shafee2024evaluation}, health-care training, customer service ~\cite{jonnala2024large}, time-series prediction, etc.

Recently, there has been increased interest in building autonomous agents using LLM, which is used as the agent's central computation engine.
To support LLM based agents, there are several supporting components such as planning, memory, and action~\cite{hua2024trustagent,wang2024survey}.

LLM-based agents can be used for a wide range of applications such as co-piloting operating systems~\cite{mei2025aiosllmagentoperating}, playing games~\cite{hu2024survey}, Web agents ~\cite{yang2024agentoccamsimplestrongbaseline}, making API calls ~\cite{du2024anytoolselfreflectivehierarchicalagents} and offering cyber security suggestions~\cite{yan2024depending}. A recent news shows a surprising use case where LLMs are used to generate lottery tickets.

Random number generation plays a crucial role in various applications, including cryptography, Web protocols such as security tokens and session identifiers, science simulations, task scheduling, resource allocations, financial asset management, optimizations, and computer games.  The quality of randomness is essential for ensuring security, fairness, and trustworthiness in these applications~\cite{bouras2024integrating}.

LLMs are not designed inherently for the generation of random numbers. Their underlying architecture is deterministic, which means that given the same input, they may produce the same output. However, recent studies have explored the potential of LLMs to exhibit randomness through techniques such as sampling from probability distributions and incorporating stochastic elements in their decision-making process~\cite{bouras2024integrating,van2024random}.

In this paper, we investigate the capability of LLMs to handle tasks that require randomness in the responses. We design a set of experiments that consider various factors that can influence the performance of an LLM in random tasks, such as types of tasks (directly or indirectly involving random sources), model states (fresh vs. non-fresh) and prompting strategies.
Our analysis includes evaluating the quality of randomness using metrics such as entropy, and comparing LLM-generated outputs to those produced by established random number generators and  algorithms.

Our research has shown essential insights into the abilities and limitations of LLMs in generating random outputs. We discovered that while LLMs can mimic randomness to a certain extent, they still struggle to achieve high quality randomness due to its inherent limitation in the algorithms and potential biases. This finding has significant implications for the scientific community, as it highlights the need for further research and development to enhance the randomness capabilities of LLMs, especially for applications where true randomness or cryptographically strong random source is critical, such as cryptography-oriented AI agents and scientific simulations (LLMs for sciences). Moreover, our research provides valuable guidance for the developers and researchers working with LLMs, enabling them to understand better and control the quality of randomness in responses of these models for various applications.


Our main contributions include:
\begin{compactitem}
    \item A benchmarking suite with multiple tasks that involve runningdonmess to test the quality of randomness generated by LLMs\footnote{Code and artifacts in GitHub. The link will be updated upon acceptance.}.
    \item A comparative analysis of LLMs and local PRNGs, highlighting the performance differences between LLMs and local methods in creating high-quality randomness.
    \item An analysis of the impact of external tools and type of tasks on an LLM’s ability to handle randomness reveals that LLMs can generate more random outputs when utilizing external tools.
    \item A discussion of potential solutions to improve the randomness of LLMs, such as incorporating entropy-based sampling and parallel chain-of-thought decoding.
\end{compactitem}

\section{Background}


\subsection{The Need for Entropy and Randomness}
GenAI and Large Language Models (LLMs) have achieved extraordinary success across a wide range of application tasks such as question answering, document summarization, decision-making, code generation, reasoning, etc. Increasingly, GenAI models have been applied to tasks that demand entropy and randomness in the responses. Emerging use cases like LLM based game engines ~\cite{wu2024instructiondrivengameengineslarge}, simulations in various domains, GenAI for sciences, LLM based API agents ~\cite{song2023restgptconnectinglargelanguage}, industrial optimizations, LLMs for making schedule decisions, all of these application tasks involve producing outputs with randomness. For example, random numbers are widely used in security protocols like challenge-response mechanisms, Web tokens, security nonce generation, session identifiers. For robustness and trustworthiness in security, it requires  GenAI based Web agents to support high quality randomness within these security protocols. In the field of scientific simulations like biology and physics, random number generators are crucial for simulating various conditions or events. The quality of entropy used in GenAI process will have direct impacts on the accuracy of the scientistic modeling and simulation outcome. For the tasks that involve scheduling and statistical sampling, for instance, randomized clinical trials, randomized resource allocations, randomized task assignments, ensuring good quality randomness is essential to avoid biases and guarantee fairness. Many emerging finance and economic use case scenarios of GenAI like randomized order execution, asset management and risk management also require high quality randomness in the outputs.

\subsection{Randomness as a Trustworthiness Problem}
While LLMs have demonstrated generating human-like text, the inherent design of token generation still poses a challenge to producing high quality random outputs to satisfy the needs of many applications. LLMs are trained on massive datasets and learn to predict the most likely next word or token based on the input and their training data. Research has shown that LLMs face challenges in generating truly random distributions. Studies have suggested that LLMs often exhibit biases and struggle to produce true randomness when it is needed, especially in complex use case scenarios. The lack of supporting reliable and good quality entropy can become a major trustworthiness issue for any application aiming to leverage LLMs for tasks where randomness is crucial to ensure fairness, reduce bias, support robustness and provide security guarantee.

\subsection{Evaluation of Randomness in LLM Responses}
With the emergence of randomness-controlled models, the effectiveness of creating entropy in the related tasks  has become a critical issue, as it may lead to biased or discriminatory outputs~\cite{huang2023bias}. 
To address these challenges, various metrics have been proposed to evaluate the randomness of LLM. Hopkins and Renda \cite{hopkins2023can} offer two sampling methods-non-autoregressive sampling (NARS) and autoregressive sampling (ARS), to evaluate the distribution sampling capabilities of LLMs in two controlled domains: uniform random number sampling and probabilistic context-free grammar (PCFG) sampling. Notably, NARS demonstrates superior performance over ARS in terms of error, variance, and containment metrics.However, Hopkins and Renda did not consider the access of external tools, like pseudo-random number generators (PRNGs), thus resulting in substantial interference to the final conclusions. Liu~\cite{liu2023does} extends this line of research by instructing GPT-4 to generate either a single number or a random sequence using varied prompts. It reveals that GPT-4 attempts to compensate for the uniformity of random numbers by sacrificing independence when functioning as a random number generator. Despite tasks directly instructing the LLMs to generate numerical values, common character-based and shuffling-related tasks should also be considered when testing the randomness of large language models.
Hence, more types of tasks involving entropy should be considered, taking into account factors such as accessibility to external tools, types of tasks, model states, and prompts. Furthermore, new metrics should be proposed to accommodate the experiments of these different tasks.

\section{Measuring LLMs' Capability on Handling Tasks Involving Randomness}\label{sec-experiment-design}

Analyzing the way that an LLM handles tasks involving randomness directly (either static or dynamic) is challenging due to the system's complexity and limited access to its internal details when it is closed source. 
Therefore, we adopt an indirect approach.
Specifically, we treat an LLM as a black box and design a series of experiments considering the major factors that may influence its performance in the tasks that involve randomness.



\subsection{Factors Considered in the Experiments Design}
We consider multiple factors in the design of our experiments to ensure a comprehensive measurement of the ability of LLMs to create randomness. These factors help to isolate the specific capabilities of LLMs and understand how different conditions might affect their performance in generating random outputs.

\paragraph{Accessibility to external tools}
LLMs can be augmented with external tools, such as pseudo-random number generators (PRNGs), to enhance their abilities in various tasks~\cite{hopkins2023can}. 
In the context of randomness, access to a PRNG can significantly influence an LLM's performance. We evaluate LLMs both with and without access to a PRNG to investigate whether they can generate random numbers independently or benefit from a traditional PRNG's assistance. This factor is crucial for understanding the inherent capabilities of LLMs in generating random numbers and whether they can achieve reasonable randomness without relying on external tools.

\paragraph{Types of tasks}
Tasks involving randomness vary significantly in complexity, the types of PRNG required (statistical PRNG, cryptographic PRNG, etc.), and how they utilize random numbers. It is often not obvious to the LLMs how randomness should be applied to meet the goals of specific tasks based on the prompt instructions. In this work, we consider two categories of tasks:

\begin{compactitem}
   \item \textit{Direct tasks.} These tasks explicitly require the generation of random numbers. Examples include generating a sequence of random integers within a specified range or producing a random floating-point number between 0 and 1. This category focuses on the LLM's ability to generate random numbers directly, a fundamental aspect of randomness in LLMs.
    \item \textit{Indirect tasks.} These tasks require an LLM to utilize random sources implicitly, often involving operations such as shuffling or sampling. Examples include shuffling a deck of cards, randomly selecting elements from a list, or generating random permutations of a sequence. This category assesses the LLMs’ ability to apply randomness in more complex scenarios where random number generation is a means to achieve a specific outcome.
\end{compactitem}

By evaluating LLMs on both the direct and indirect tasks, we aim to assess their ability to handle randomness across a spectrum of applications. 

\paragraph{Model states}
The internal state of an LLM, influenced by its previous interaction history, can affect its behavior on new tasks. To account for this, we consider both ``fresh'' (newly initialized) and ``non-fresh'' (previously used) LLM models in our experiments. 
This distinction allows us to investigate whether prior experiences influence an LLM's ability to generate random outputs. 

Interestingly, research suggests that LLMs might not only inherit human biases in randomness generation but could potentially amplify them \cite{van2024random}. 
This highlights the challenges of achieving true randomness in LLMs and the need for careful evaluation.

\paragraph{Prompts}
Prompts are the only source for an LLM to receive external task descriptions in the inference phase.
Many works have demonstrated that carefully designed prompts can lead to better LLM responses~\cite{sahoo2024systematic}.
For the same task involving randomness, it is very likely that different prompts result in different randomness handling. 


\subsection{Experiment Categories}

Based on the factors outlined above, we systematically evaluate task types on an LLM's ability to handle randomness, while also considering the influences of model status.
To ensure a comprehensive evaluation, we included a variety of LLMs in our study, namely OpenAI's GPT-4o, Google's Gemini 1.5 Pro, Mistral Large(2047), Gemma2 27b and Llama 3.1 8b. This selection allows us to evaluate randomness generation capabilities across different models and architectures.

To rigorously evaluate the capability of LLMs to handle tasks that involve randomness and generate the output based on the given task, we designed a series of experiments encompassing three distinct task categories: numerical, character-based, and shuffling-related. These categories were selected to represent different types of applications for randomness, from basic number generation to more complex tasks involving sequences and permutations. This methodology draws inspiration from the existing research on evaluating LLM sampling in controlled domains~\cite{hopkins2023can} and aims to provide a comprehensive assessment of LLMs' ability to exhibit randomness across various scenarios. 


\subsection{Numerical Value Related Tasks}

Random number generation is widely recognized as the most fundamental task associated with randomness.  Our evaluation considers LLM-based agents with and without access to a pseudo-random number generator (PRNG). This allows us to investigate whether LLMs can generate random numbers independently or benefit from a traditional PRNG's assistance, similar to the approach used in the previous studies ~\cite{bouras2024integrating}. Furthermore, we vary the scale of the task, including the size of the output (e.g., generating a single number versus a sequence of numbers) and the range of random numbers (e.g., integers within a specific interval, floating point values). This manipulation of scale allows us to assess the impact of these factors on the LLM's performance and identify potential limitations in its ability to generate random numbers across different magnitudes and data types. 

We employ statistical tests commonly used to evaluate random number generators, such as the well-defined tests in the NIST randomness test suite ~\cite{rand}, to assess the quality of the generated random numbers. These tests will help us determine if the distribution of the LLM-generated numbers significantly deviates from a truly uniform distribution, indicating potential biases or patterns in the LLM's outputs.

\subsection{Characters Related Tasks}

This category requires the LLM-based agent to manipulate characters randomly, such as generating random strings given a specified alphabet (e.g., generating random passwords, or random sequences of letters from the English alphabet). While this task can theoretically be reduced to random number generation by mapping characters to numerical values, it remains unclear whether LLMs can effectively leverage this relationship. Our experiments aim to determine whether LLMs can exhibit randomness in character-based tasks without explicit reliance on numerical methods.

To evaluate the randomness of the generated character strings such as passwords, we analyze their statistical properties, such as the frequency distribution of individual characters and the presence of recurring patterns or sub-strings. 

\subsection{Shuffling Related Tasks}

Shuffling is a classic application of randomness with extensive research in various domains, including card games, data analysis, and algorithm design. The problem of card shuffling, while seemingly a simple use case, has been intensively studied in the field of applied statistics, not only due to its significance to ensure fairness in card games, but also its applicability in diverse fields, ranging from game design, data analysis to scientific research. 

In card shuffling, a well-shuffled deck ensures that each player has an equal chance of receiving any particular card or combination of cards. A strong uniform stopping time ensures that after the stop, the deck is in a truly random state, regardless of how it was initially arranged. Theoretical results have been obtained to understand stopping times in the context of conventional card shuffling techniques ~\cite{DIACONIS1983175,aldous_shuffling_1986,fa7e9dd1-a4a5-3f18-9c68-46611747ec81}. 

In this study, we evaluate LLMs on tasks analogous to shuffling cards, such as assigning patients to doctors or permuting a set of words. These tasks represent a higher complexity level than simple number or character generation, requiring the LLMs to understand and apply the concept of random permutations.

We analyze the generated permutations for uniformity and randomness to assess the LLMs’ performance on shuffling tasks. We compare the LLM-generated permutations to those produced by a well-established shuffling algorithm to identify any biases or deviations from the expected randomness. 

By conducting these experiments across a range of tasks and evaluating the results using appropriate statistical measures, we aim to provide a systematic assessment of LLMs' capacity for generating random output and identify potential areas for improvement in their ability to exhibit true randomness. The scoring method for the shuffle process is based on the minimum entropy of the distances between each pair of cards after the shuffle is applied. This effectively captures the idea that shuffling should maximize statistical entropy, which is a measure of randomness that can be calculated when the inner workings of the random system are unknown.



Below is how we compute statistical entropy as a percentage of the maximum possible entropy:
\begin{enumerate}
\small
    \item N := number of cards in the deck
    \item \(K_{ijd}\) := after how many trials card i was distance d from card j
    \item H := percent of max entropy, 
    
    min\(\{-\sum_{d=1}^{N-1}(\frac{K_{ijd}}{\sum{K_{ij}}}log_{N-1}\frac{k_{ijd}}{\sum K{ij}}) | 0<i, j<N\}\)
\end{enumerate}
The above method has been used to evaluate the quality of different card shuffling algorithms. 


\subsection{Direct Generation}

We evaluate three specific domains to assess whether the LLMs can generate randomness in their responses. Our goal, however, was not only to deal with randomness generation, but also to investigate how the LLMs handle tasks that require randomness as part of their core function but were not asked to generate it directly like random number sequence creation.

We instructed the LLMs to generate random numbers and then apply the NIST randomness test-suite to evaluate whether they could pass the randomness tests. 
Each LLM was prompted to generate 10000 random 8-bit integers. The prompts used were as follows:
\begin{itemize}
\item \textbf{GPT-4o} "Can you please create 10000 random positive integers in decimal format, the highest of which is 255."
\item \textbf{Gemini and Others} "Can you please create 10000 random positive integers in decimal format, the highest of which is 255." with System Prompt "You are a true random number generator. You will be asked to generate random numbers in JSON format. Do not give any code or ideas. Only the answer."
\end{itemize}

The generated sequences were then subjected to the Random Number Test Suite~\cite{rand} to assess the quality of randomness ~\cite{van2024random}.

\subsection{Evaluation Metrics}

We utilized the Random Test Tool (RTT)~\cite{githubRandom_test_toolREADMEmdMain} to assess the quality of randomness in the number of sequences generated by the LLMs. This tool provides a comprehensive suite of statistical tests designed to evaluate various aspects of randomness, including uniformity, independence, and unpredictability. 


The tool implements the popular NIST Randomness Test Suite~\cite{10.5555/2206233}. Compared with similar tools, RTT is user-friendly, easy to manipulate, and capable of delivering clear results to users. The set of NIST Tests supported are.

\textbf{Monobit (Chi2)} This test is intended to see if the frequencies of 1 and 0 across the entire n-bit sequence are approximately equal, meaning that the proportion of 1s and 0s is close to half. If the number of 0s and 1s are not the same, it is intended to see if their difference falls within the limit of randomness. 

\textbf{Frequency in block} This test is intended to ensure that frequencies of 1 and 0 are evenly distributed across the entire n-bit sequence.

\textbf{Run Test}  This test is intended to see if the frequencies of runs of 1s and 0s of various
lengths would be within the limits of randomness.

\textbf{Longest run of Ones} This test is intended to see if the frequencies of the longest run of 1s of various lengths appearing in the sequence are consistent with that expected for a random sequence.

\textbf{Binary Rank} This test is intended to see if the n-bit string has repetitive patterns across its entire sequence. The n-bit string is sequentially divided into N disjoint blocks, and it endeavors to see linear dependence among its fixed length substrings of each block.

\textbf{Linear Complexity} A long-bit string is usually obtained from an LFSR (Linear Feedback Shift Register). The bit sequence from which a longer LFSR is obtained can be termed as random, while the shorter LFSR indicates non-randomness. The linear complexity test looks for the length of the LFSR and determines if the bit sequence from which the LFSR is obtained is random or not. 

\textbf{Serial Test} The serial test counts the frequency of all possible overlapping m-bit patterns across the entire n-bit sequence, and based on the deviations of each of the counts together, one intends to see if the sequence can be termed as random or not.

\textbf{Spectral} The focus of this test is the peak heights in the Discrete Fourier Transform of the sequence. The purpose of this test is to detect periodic features (i.e., repetitive patterns that are near each other) in the tested sequence that would indicate a deviation from the assumption of randomness. 

\textbf{Sign} This test checks the equal repartition of the data around the median.

Given a random sequence, the tests calculate a \textit{p-value}. The \textit{p-value} represents the probability of obtaining a distribution at least as extreme as that observed. Then, the tool compares the \textit{p-value} to a threshold (0.01). If the \textit{p-value} is lower than this threshold, it implies that the probability of the observed behavior occurring by chance is 1 - 0.01 (99\%). For \textit{p-value} $< 0.01$, a perfectly random sample is expected to fail this test only 1\% of the time. The tool follows the default classification:

\begin{table}[]
\centering
\caption{\textit{p-value} Interpretation}
\label{tab-p-value-interpretation}
\begin{tabular}{p{1.2cm}|p{3.8cm}p{2.3cm}}
\hline
\textbf{Label} & \textbf{\textit{p-value}} & \textbf{Interpretation} \\                   
\hline
OK & 0.1  \textless  \textit{p} \textless 0.99  & Test successful \\
SUSPECT  &  0.01 \textless  \textit{p} \textless  0.1 or 0.9 \textless \textit{p} \textless 0.99   &  Suggest to re-test \\
KO  & \textit{p} \textless 0.01 and \textit{p} \textgreater 0.99 &   Test failure  \\
\hline
\end{tabular}
\end{table}


Interpretation of p-values and classification of results adhere to the guidelines provided in the RTT documentation \cite{rand}.
\begin{itemize}
    \item \textbf{OK:} The test is successful if the p-value falls within the range [0.1, 0.99]. This indicates that the observed distribution is consistent with what would be expected from a truly random sequence.
    \item \textbf{SUSPECT:} If the p-value is in the intervals [0.01, 0.1] or [0.9, 0.99], it suggests a potential deviation from randomness. Further investigation and re-testing are necessary to confirm or reject the null hypothesis of randomness.
    \item \textbf{KO:} The test fails if the p-value is $< 0.01$ or $> 0.99$. This indicates a statistically significant deviation from randomness, suggesting that the sequence is likely not random. 
\end{itemize}

\subsection{Remark on Reproducibility}
Reproducibility is a crucial property that facilitates trust in certain applications of GenAI, for instance, healthcare use cases, finance, legal applications, many scenarios in various science domains. In case that GenAI responses to application tasks depend on random sampling, and entropy, reproducibility should be interpreted within the context of the tasks. For instance, for card shuffling, it makes more sense to define reproducibility as producing similar random distribution of the shuffled cards after performing the same of number of card shuffles in each trial. For the tasks relying on random sampling, 
reproducibility needs to be defined as reproducible distribution of the outputs given the same experiment setting. This is the interpretation of reproducibility that we have taken in the context of this endeavor. During experiments, we conducted data collection in multiple iterations to ensure consistent entropy measurements across different trials.

\section{Experimental Results and Analysis}\label{sec-results-analysis}


For our experiments, we choose a combination of open-weight and closed-weight models. The open models allow us to dig more deeply into their generation strategies and allow us to play with the temperatures. Whereas for the closed models, we used both an official API and a web-based interface to collect data.

\subsection{Analysis of NIST Randomness Tests}

This analysis compares the randomness quality of various Large Language Models (LLMs) and local Python random number generators based on the results from the NIST randomness testsuit.

\begin{table}[h]
\centering
\caption{Percentage of test outcomes for different random number generators}
\label{tab:random_generators}
\scriptsize
\resizebox{\columnwidth}{!}{%
\begin{tabular}{|l|c|c|c|}
\hline
\textbf{Generation Methods and LLM} & \textbf{OK} & \textbf{SUSPECT} & \textbf{KO} \\ \hline
Local 8-bit numbers                 & 87.78\%     & 11.11\%          & 1.11\%      \\
Local 1M data sample                & 91.11\%     & 7.78\%           & 1.11\%      \\
random.SystemRandom()               & 87.78\%     & 11.11\%          & 1.11\%      \\
secrets.randbelow()                 & 88.89\%     & 8.89\%           & 2.22\%      \\
Gemini 15                           & 30.56\%     & 13.89\%          & 55.56\%     \\
Phi-3 (Run 1)                       & 22.22\%     & 0\%              & 77.78\%     \\
Phi-3 (Run 2)                       & 25\%        & 12.5\%           & 62.5\%      \\
Gemma 2 27B                         & 11.11\%     & 0\%              & 88.89\%     \\ \hline
\end{tabular}%
}
\end{table}

\subsubsection{Key Observations}

\begin{enumerate}
    \item \textbf{Local Python Generators:} The local Python random number generators (8-bit numbers, 1M data sample, SystemRandom, and secrets.randbelow) performed significantly better than the LLMs, with over 87\% of tests passing (OK) for all four methods.
    
    \item \textbf{LLM Performance:} The LLMs (Gemini 15, Phi-3, and Gemma 2 27B) showed poor randomness qualities, with a high percentage of failed (KO) tests.
    
    \item \textbf{Gemini 15:} Performed the best among the evaluated LLMs, passing 30.56\% of tests, but still significantly underperformed compared to the local Python generators.
    
    \item \textbf{Phi-3:} Showed inconsistent results between two runs, with Run 2 performing slightly better than Run 1.
    
    \item \textbf{Gemma 2 27B:} Demonstrated the poorest performance among the measurable models, failing 88.89\% of the tests.
    
    
    \item \textbf{Honorable Mention:} We tried to run these tests with llama 3.1 8b model as well. That did not pass any test.
\end{enumerate}

The analysis reveals a clear distinction between the randomness quality of the local Python random number generators and LLMs. Local Python methods consistently produced high-quality random numbers, passing most statistical tests. In contrast, the LLMs struggled to generate truly random sequences, with Gemma 2 27B performing particularly poorly.

These results suggest that LLMs, in their current state, are not suitable for applications requiring high-quality random number generation. 

\subsubsection{Qualitative Analysis}

While the quantitative results provide a statistical assessment of randomness, a qualitative analysis can reveal further insights. Observing the generated sequences, we noticed certain patterns and tendencies that deviate from ideal randomness. For example, some LLMs exhibited a slight bias towards generating certain numbers or ranges of numbers more frequently than others. This observation aligns with the previous research that highlights the limitations of LLMs in accurately mimicking target distributions~\cite{hopkins2023can}. This is specially relevant when talking about the LLMs that try to generate random numbers on their own without relying on external tools. 

\begin{figure}
    \centering
    \includegraphics[width=0.75\linewidth,height=1.0in]{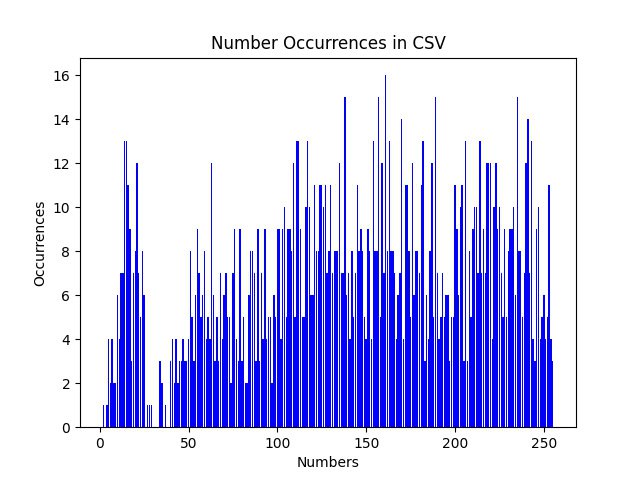}
    \caption{Gemini 1.5 Pro distribution of random numbers.}
    \label{fig:gemini10k}
\end{figure}
\begin{figure}
    \centering
    \includegraphics[width=0.75\linewidth,height=1.0in]{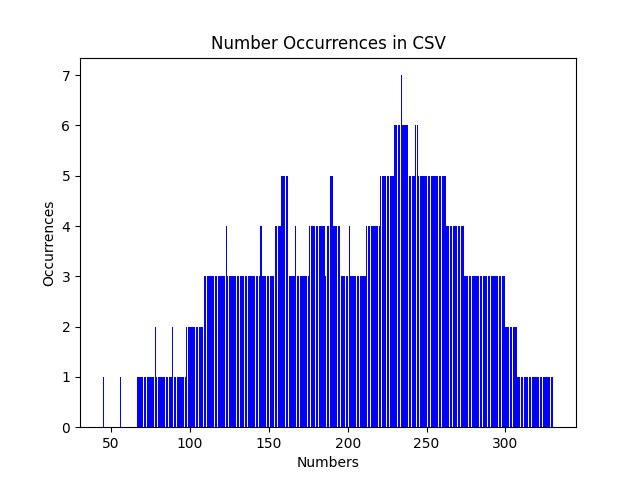}
    \caption{Mistral Large(2047) distribution of random numbers.}
    \label{fig:mistral10k}
\end{figure}
\begin{figure}
    \centering
    \includegraphics[width=0.75\linewidth,height=1.0in]{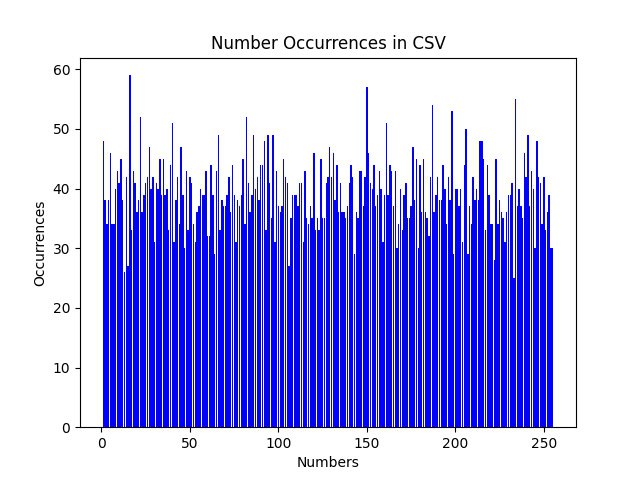}
    \caption{Gemini 1.5 with function calling (10k).}
    \label{fig:geminitool10k}
\end{figure}

Figure \ref{fig:gemini10k} and Figure \ref{fig:mistral10k} show the distribution of random numbers between 0 to 255 when generated 10000 times. These generations were done using the described method in Section \ref{sec-experiment-design} and do not use any function call.

Even without the randomness tests, we can see the distribution is not random, and each model favors a specific set of numbers. For Gemini 1.5 Pro Fig: \ref{fig:gemini10k}, the top 3 Preferred Numbers are 161 (16 occurrences),138 (15 occurrences), and 235 (15 occurrences). For Mistral Fig:\ref{fig:mistral10k}, this behavior is even more clustered, with the cluster being in the middle with 234 (7 occurrences), 243 (6 occurrences), and 230 (also 6 occurrences).

\begin{figure}
    \centering
    \includegraphics[width=0.7\linewidth]{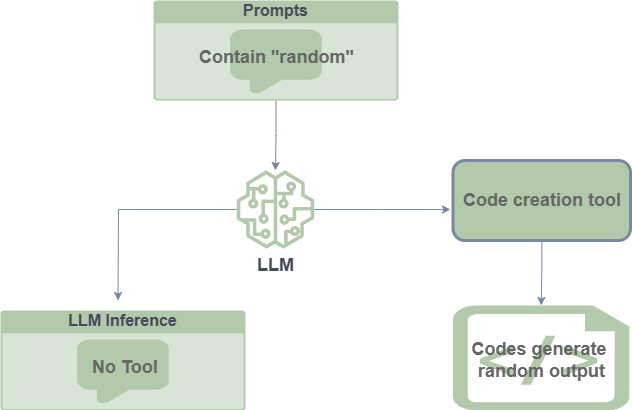}
    \caption{LLM function call workflow.}
    \label{fig:llmcall}
\end{figure}

\begin{figure}
    \centering
    \includegraphics[width=0.75\linewidth,height=1.5in]{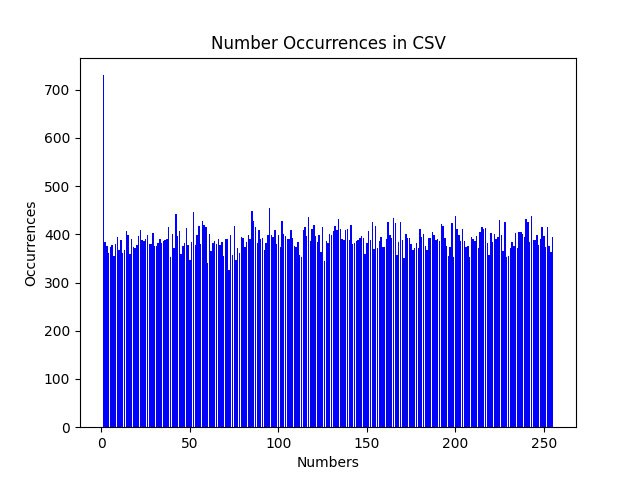}
    \caption{Gemini 1.5 with function calling (100k).}
    \label{fig:geminitool100k}
\end{figure}

However, when we have Gemini with function calling enabled, we see in Fig \ref{fig:geminitool10k} the numbers being almost evenly distributed for 10,000 random number generation. This is where we observe that the model is defaulting to the workflow as depicted in Figure \ref{fig:llmcall}. Things take a bit different turn when we have 100,000 random numbers generated with Gemini using function calling, as we see in Figure \ref{fig:geminitool100k} where almost the whole spectrum of numbers is evenly distributed apart from Number 1, which disproportionately is chosen 730 times.



\subsection{Shuffling Results}
This section analyzes the results of the shuffle method obtained through our experimental process.

\subsubsection{Method}
We focus on evaluating the effectiveness of different shuffling methods and quantifying the degree of randomness achieved using an entropy-based metric. We aim to gain insights into the factors influencing randomness generation in these models by analyzing the entropy values obtained from various LLMs and shuffling scenarios.

To assess the randomness generation capabilities of LLMs, we designed experiments centered around shuffling a deck of playing cards. This task provides a well-defined and quantifiable measure of randomness, allowing for a systematic evaluation of different LLMs and shuffling methods. We employed two primary shuffling scenarios: (1) Local shuffle, where the LLM generates Python code executed to shuffle the deck, and (2) LLM shuffle, where the LLMs directly shuffle the deck without external code.

To quantify the degree of randomness achieved in the shuffled decks, we utilized the entropy measure described earlier. Our entropy definition grades the quality of the unknown shuffling method by measuring pairwise distances between two cards. In the context of card shuffling, higher entropy values correspond to greater randomness in the sequence of cards. 
We varied the number of shuffle rounds (128 to 2048) and recorded the corresponding entropy values for all the tested scenarios.

We conducted experiments with several LLMs, including variants of GPT, Gemini, and Llama. For each LLM, we systematically varied the number of shuffle rounds (from 128 to 2048) in both Local and LLM shuffle scenarios. This allowed us to observe how the degree of randomness changed with increased rounds of shuffling, and identify potential convergence patterns.

Table \ref{tab:card-shuffling-entropy} summarizes the entropy results for the card shuffling experiments conducted with GPT, Gemini, and the local shuffle scenarios.

\begin{table}
\centering
\caption{Entropy results for the card shuffling experiments.}
\label{tab:card-shuffling-entropy}
\scriptsize
\resizebox{\columnwidth}{!}{%
\begin{tabular}{|l|l|l|l|}
\hline
    Shuffle rounds & Local             & GPT               & Gemini             \\ \hline
    128            & 0.874451358681846 & 0.861846152487087 & 0.87337156361593   \\ \hline
    256            & 0.897806026358848 & 0.876740097767342 & 0.899403147801315  \\ \hline
    512            & 0.899562541724958 & 0.895595748483194 & 0.908432001491059  \\ \hline
    786            & 0.916425927773246 & 0.905871556497765 & 0.910063521198634  \\ \hline
    1024           & 0.915989771924721 & 0.909819923462218 & 0.912599250807944  \\ \hline
    1280           & 0.914072526105781 & 0.915665610283595 & 0.913379487780489  \\ \hline
    1536           & 0.918177419740495 & 0.918220137760354 & 0.917924657467474  \\ \hline
    1792           & 0.921352692988455 & 0.921004938696014 & 0.923022991807562  \\ \hline
    2048           & 0.923277055991329 & 0.924444147944848 & 0.92356798431639  \\ \hline
\end{tabular}
}

\end{table}

As the number of shuffle rounds increases, the entropy values for the local, GPT, and Gemini shuffles exhibit a clear trend of convergence toward a high entropy value. The entropy values for all three are remarkably close across all the shuffle rounds. This suggests that the LLMs demonstrate a comparable ability to generate random shuffles given sufficient number of shuffling rounds.

To further investigate the generalizability of our findings, we conducted a card shuffling experiment using the Llama LLM. Table \ref{tab:card_entropy_llama} presents the entropy results for Llama using the GPT shuffle scenario.
\begin{table}[h]
\caption{Entropy results of Llama card shuffles. }
\label{tab:card_entropy_llama}
\centering
\scriptsize
\resizebox{\columnwidth}{!}{%
\begin{tabular}{l|llll}
\hline
\textbf{Rounds}  & 128       & 256       & 512      & 1024      \\
\textbf{Entropy} & 0.7869377 & 0.7885018 & 0.805581 & 0.8087705 \\ \hline
\end{tabular}%
}
\end{table}

The Llama 3.1 8b exhibits consistently lower entropy values compared to GPT and the local shuffles, suggesting potential limitations in its ability to generate random sequences. This discrepancy could be attributed to differences in model architecture, training data, or specific prompting techniques used.

Using the defined entropy measurement, we compare the entropy of GPT card shuffles with the result of the locally shuffled cards. To reduce the token count, we asked GPT to shuffle ten cards up to 2048 times (rounds). Local card shuffles were performed using the Python card shuffle code returned from GPT.  The entropy results indicate that there is almost no difference between the result of GPT and the result of local shuffles.   



It is often assumed that adjusting the temperature parameter in LLMs can significantly improve the randomness of their outputs. However, our experiments suggest that this might not always be the case. For specific tasks, tweaking the temperature settings may not lead to substantial changes in the underlying concepts generated by the LLM, even if the specific words used vary. This observation highlights the need for alternative approaches to enhance randomness in LLMs, potentially focusing on refining prompting techniques, incorporating stochastic elements in the model architecture, or exploring different sampling methods.

\subsection{Analysis of LLM Generated Passwords}

This analysis compares the randomness quality of the password sequences created by the LLMs. In these
test scenarios, the LLMs were instructed to generate random passwords of a certain length with characters chosen from the English alphabet (both the lower and uppercase letters), decimal digits, and a set of special characters. Randomness of the generated passwords were tested with the same NIST randomness test tools. 

\begin{table}[h]
\centering
\caption{Percentage of test outcomes for LLM generated password sequences.}
\label{tab:random_pw}
\scriptsize
\resizebox{\columnwidth}{!}{%
\begin{tabular}{|l|c|c|c|}
\hline
\textbf{Generation Methods and LLM} & \textbf{OK} & \textbf{SUSPECT} & \textbf{KO} \\ \hline
GPT   &  44.4\% & 11.1\%, & 44.5\% \\
Gemini 15                           & 33.3\%     & 11.1\%          & 55.6\%     \\
Phi-3                       & 0\%     & 11.1\%              & 88.9\%     \\
Gemma 2 27B                         & 0\%     & 0\%              & 100\%     \\ \hline
\end{tabular}%
}
\end{table}

The results indicate that the password sequences generated by the LLMs exhibit poor entropy and low quality randomness. The percentage of the failed NIST randomness tests was high for all the evaluated LLMs. Examining the generated passwords by Gemma 2 and Phi-3 show that the generated password sequences contained repeated passwords, or repeated substrings. 

\subsection{Summary of Factors Affecting LLMs Capability on Handling of Randomness}

While LLMs can exhibit variability in their outputs, achieving true randomness remains challenging due to their deterministic training process and inherent biases\cite{AIsDicey}. These models are trained on massive datasets, learning to predict the next word in a sequence based on patterns and relationships within the data\cite{HowLarge}. While enabling impressive language generation capabilities, this process can limit their ability to produce truly random outputs ~\cite{van2024random}. 

Several factors contribute to LLMs limitations in handling tasks involve randomness:
\begin{compactitem}
    \item \textbf{Bias in the training data}: If the training data contains biases or predominantly reflects specific patterns, the model's output may be skewed, even when randomness is desired~\cite{BiasRandomnessandRi}.
    \item \textbf{Limited knowledge update}: LLMs cannot update their knowledge base in real-time, hindering their ability to incorporate new information and adapt to changing contexts, which is crucial for generating random outputs based on the latest information~\cite{10Bigge}.
    \item \textbf{Lack of true understanding}: LLMs do not possess genuine comprehension of the text they generate, which can lead to non-sensical or irrelevant output, especially when dealing with complex or nuanced concepts that require randomness~\cite{AllYouNeedtoKnowab}.
\end{compactitem}

Despite these limitations, LLMs have shown surprising capabilities in certain scenarios, such as generating numbers from specific distributions without explicit programming. This suggests potential for improvement and further research to enhance their handling of randomness.  

Potential solutions include improving the diversity of training data ~\cite{bouras2024integrating}, developing new algorithms to generate random sequences ~\cite{HowtoGetBetterOutputs}, incorporating feedback mechanisms to refine the performance of the model ~\cite{LLMChallenges}, and fine-tuning task mixtures to improve generalization. Additionally, embracing and leveraging the inherent randomness of LLMs, particularly in creative applications, can be a viable approach ~\cite{LLMsDeterminismRandomness}. 

Furthermore, controlling randomness with parameters such as temperature and top p allows users to fine-tune the balance between predictability and variability in LLM outputs ~\cite{LLMsDeterminismRandomness}. These parameters provide a degree of control over the generation process, enabling users to influence the randomness and creativity of the model's responses.

\subsection{Ability to Follow Prompt Instructions}
During experimentation, we made the following observations. Firstly, for some models, particularly those with small size, the ability to precisely follow prompt instructions is sometimes poor compared to larger models.  For example, the output may not always be in the format requested by the prompt instructions. To fix this, in certain cases, it requires post-processing of the outputs to convert them into format that can be processed by the randomness analysis tools. Secondly, when not emphasized in the prompts, LLMs may output programming code for the requested task instead of producing the outputs based on the inputs. For example, when asked to shuffle cards, a LLM may give a card shuffle Python code as response. We mitigated this issue with more specific prompts to indicate that the output should not be code. Thirdly, some models occasionally do not perform the requested number of randomization steps. For instance, when asked to shuffle cards 1,000 times, a model may shuffle only 50 times. Since we repeated data collection with many trials and computed entropy over a large number of outputs across trials, this has been less than a problem because we can resume from where it left off.

\section{Reproducibility Framework}
The stochastic nature of Large Language Models (LLMs) presents significant challenges to the scientific principle of reproducibility. To address this, we established a rigorous framework to ensure the experiments presented in this paper can be reliably and accurately replicated, combining methodological constraints, task-specific interpretations of reproducibility, and a commitment to open artifact availability. A primary source of non-determinism in LLMs is the sampling strategy; therefore, to create a consistent baseline, all interactions were conducted with the \textbf{temperature parameter set to 0}. This greedy decoding approach makes the generation process deterministic for a given prompt and model state, which is essential for evaluating core capabilities. We posit that reproducibility for tasks involving randomness is achieved not through identical outputs, but through consistent statistical properties. Accordingly, for direct generation tasks like random numbers and strings, reproducibility is assessed by applying the NIST Statistical Test Suite (\texttt{sts-2.1.2}) \cite{rukhin2001statistical} to generated sequences, where a successful replication involves passing or failing the same statistical tests. For indirect tasks like card shuffling, reproducibility is defined by the convergence of statistical entropy, using a metric that measures pairwise distances between items (detailed in Section 3.5), with successful replication indicated by convergence to values statistically indistinguishable from those in Table 3 and Table 4. 


\section{Limitations}
Our study has several limitations that should be acknowledged. First, the evaluation is based on a specific set of LLMs and tasks, and the results may not be generalizable to other LLMs or tasks.  
Another limitation is the inherent nature of transformer architectures. We run all our experiments with temperature 0 to facilitate reproducibility. However, that also introduces a certain determinism when generating random numbers by prompting. That makes our results and work, dependent on prompting. 
Finally, 
since our evaluation includes closed-weight models,
the lack of access to the internal workings of these models makes it difficult to fully understand and control the factors that influence their randomness generation capabilities.  Future work will explore the use of open-weight models and more transparent evaluation methods and provide deeper insights into the mechanisms underlying randomness in LLMs.

\section{Related Works}\label{sec-related}

The evaluation of randomness generation in LLMs is an emerging field of study. Hopkins and Renda [14] provided one of the first empirical evaluations of LLMs as distribution samplers, establishing key metrics and highlighting performance differences between autoregressive and non-autoregressive sampling. Liu [19] extended this by focusing on GPT-4's ability to generate random numerical sequences, revealing that the model often compensates for uniformity by sacrificing independence. More recently, Harrison [13] compared LLM and human performance on random number generation, finding that models may still not match human-level capabilities in this specific task. Our work builds directly on these foundational studies by providing a more comprehensive evaluation framework that includes a wider variety of direct and indirect randomness tasks (numerical, character, and shuffling), applies the full NIST suite of randomness tests for a more rigorous assessment, and analyzes a broader set of contemporary LLMs.

Our focus on a rigorous reproducibility framework also situates this paper within the broader conversation on the challenges of reproducibility in machine learning research. The difficulty of replicating results in computational science, often termed the "reproducibility crisis," is particularly acute in deep learning due to numerous sources of non-determinism \cite{gundersen2018state}. Even with fixed random seeds, subtle variations in software libraries (e.g., cuDNN versions), hardware (e.g., GPU architecture), and the non-deterministic nature of certain parallelized floating-point operations can lead to divergent outcomes \cite{sculley2015hidden}.

In response, the machine learning community has proposed various best practices to mitigate these issues. Pineau et al. \cite{pineau2021improving} introduced a widely recognized checklist for reproducibility, encouraging the publication of not only code but also model weights, hyperparameters, and detailed execution environments. The use of containerization technologies like Docker has also been advocated as a method to encapsulate the full software stack, ensuring that dependencies and system configurations can be perfectly replicated \cite{gundersen2018state}. Our work contributes to this effort by not only adhering to these principles through the public release of our code and experiment artifacts but also by proposing a task-specific definition of reproducibility for stochastic LLM evaluations, where statistical consistency, rather than identical output, serves as the benchmark for a successful replication.

\section{Ethical Implications and Responsible Disclosure}

The findings of this study, which highlight significant deficiencies in the ability of Large Language Models (LLMs) to generate high-quality randomness, carry substantial ethical implications. As LLMs are increasingly integrated into a wide array of applications, from consumer-facing tools to enterprise-level systems, a misunderstanding of their limitations in stochastic processes could lead to predictable, insecure, and biased outcomes. The demonstrated weakness in generating random passwords, for instance, poses a direct security risk. If developers or end-users mistakenly trust an LLM to generate cryptographic material or unique identifiers, the resulting outputs could be vulnerable to adversarial prediction and compromise system security. This aligns with concerns raised by Bourtoule et al. \cite{bourtoule2021machine} regarding the unforeseen failure modes of machine learning systems when deployed in critical environments.

Furthermore, the tendency of LLMs to exhibit biases, as noted in Section 4.4, can be amplified when randomness is expected but not properly delivered. For example, in a system designed to randomly assign resources or opportunities (e.g., in randomized clinical trials or automated scheduling), a biased "random" process could lead to systematically unfair or inequitable outcomes, perpetuating societal biases present in the training data \cite{mehrabi2021survey}. This underscores the ethical imperative for developers and researchers to be transparent about the capabilities and limitations of their models.

Consequently, we have a responsibility to disclose these findings in a manner that informs without causing undue alarm or enabling malicious actors. Our approach to responsible disclosure involves two key actions. First, by publishing this research in a peer-reviewed venue, we aim to alert the academic and industrial communities to these potential vulnerabilities, encouraging the development of more robust systems and best practices. Second, we advocate for clear guidelines and warnings within developer documentation for LLMs, explicitly cautioning against their use as a primary source of entropy for security-sensitive or fairness-critical applications. This aligns with the principle of "transparency" in AI ethics, which calls for clear communication about how AI systems operate and where they might fail \cite{floridi2019establishing}. We believe that a proactive and transparent approach is the most effective way to mitigate the risks associated with the misuse of LLMs in contexts requiring true or high-quality randomness.

\section{Conclusions}

To gain a better understanding of LLM-based agents' capabilities in handling tasks that involve randomness, we developed a set of experiments and tested several LLMs.  Our analysis included evaluating the quality of randomness using metrics like entropy and well established NIST randomness test-suite. The results show that while LLMs can mimic randomness to a certain extent, they still struggle to achieve high quality randomness.
This study contributes valuable insights into the capabilities and limitations of LLMs in generating random outputs. 


\section*{Acknowledgment}
This research was supported in part by the Google Developer Experts program, which provided Google Cloud research credits to enable our experiments.\footnote{This material is based upon work supported by the Google Cloud Research Credits program.} Additionally, we gratefully acknowledge the financial and collaborative support of the NATO Science for Peace and Security (SPS) Programme\footnote{See \url{https://www.nato.int/cps/en/natohq/78209.htm}}, which fosters international cooperation in scientific research.

\bibliographystyle{plainnat}
\bibliography{dekhi}
\end{document}